\documentclass[sn-mathphys-num]{sn-jnl}

\usepackage{graphicx}%
\usepackage{multirow}%
\usepackage{amsmath,amssymb,amsfonts}%
\usepackage{amsthm}%
\usepackage{mathrsfs}%
\usepackage[title]{appendix}%
\usepackage{textcomp}%
\usepackage{manyfoot}%
\usepackage{booktabs}%
\usepackage{algorithm}%
\usepackage{algorithmicx}%
\usepackage{algpseudocode}%
\usepackage{listings}%
\usepackage{subcaption}%
\usepackage[T1]{fontenc}
\usepackage{lmodern}
\usepackage[table,xcdraw]{xcolor}
\usepackage{hhline}
\usepackage{tabularx} 
\usepackage{adjustbox} 
\usepackage{makecell}
\usepackage{wrapfig}

\begin{document}

\title[Article Title]{FocalAD: Local Motion Planning for End-to-End Autonomous Driving}


\author[1]{\fnm{Bin} \sur{Sun}}

\author[1]{\fnm{Boao} \sur{Zhang}}
\author[1]{\fnm{Jiayi} \sur{Lu}}
\author[1]{\fnm{Xinjie} \sur{Feng}}
\author[1]{\fnm{Jiachen} \sur{Shang}}
\author[1]{\fnm{Rui} \sur{Cao}}
\author[1]{\fnm{Mengchao} \sur{Zheng}}
\author[1]{\fnm{Chuanye} \sur{Wang}}
\author*[1]{\fnm{Shichun} \sur{Yang}}
\email{yangshichun@buaa.edu.cn}
\author*[2,3]{\fnm{Yaoguang} \sur{Cao}}
\email{caoyaoguang@buaa.edu.cn}
\author*[4]{\fnm{Ziying} \sur{Song}}
\email{songziying@bjtu.edu.cn}

\affil[1]{\orgdiv{School of Transportation Science and Engineering}, \orgname{Beihang University}}

\affil[2]{\orgdiv{State Key Lab of Intelligent Transportation System}, \orgaddress{\city{Beijing},  \country{China}}}

\affil[3]{\orgdiv{Hangzhou International Innovation Institute}, \orgname{Beihang University}}

\affil[4]{\vspace{-20pt}\orgdiv{School of Computer Science and Technolog}, \orgname{Beijing Jiaotong University}}


\abstract{In end-to-end autonomous driving,the motion prediction plays a pivotal role in ego-vehicle planning. However, existing methods often rely on globally aggregated motion features, ignoring the fact that planning decisions are primarily influenced by a small number of locally interacting agents. Failing to attend to these critical local interactions can obscure potential risks and undermine planning reliability. In this work, we propose \textbf{FocalAD}, a novel end-to-end autonomous driving framework that focuses on critical local neighbors and refines planning by enhancing local motion representations. Specifically, FocalAD comprises two core modules: the \textbf{Ego-Local-Agents Interactor (ELAI)} and the \textbf{Focal-Local-Agents Loss (FLA Loss)}. ELAI conducts a graph-based ego-centric interaction representation that captures motion dynamics with local neighbors to enhance both ego planning and agent motion queries. FLA Loss increases the weights of decision-critical neighboring agents, guiding the model to prioritize those more relevant to planning. Extensive experiments show that FocalAD outperforms existing state-of-the-art methods on the open-loop nuScenes datasets and closed-loop Bench2Drive benchmark. Notably, on the robustness-focused Adv-nuScenes dataset, FocalAD achieves even greater improvements, reducing the average collision rate by\textbf{ 41.9\%} compared to DiffusionDrive and by \textbf{15.6\%} compared to SparseDrive.}




\maketitle

\section{Introduction}\label{sec1}

\begin{figure}[htbp]
    \centering
    \begin{subfigure}{0.495\linewidth}
        \centering
        \includegraphics[width=\linewidth]{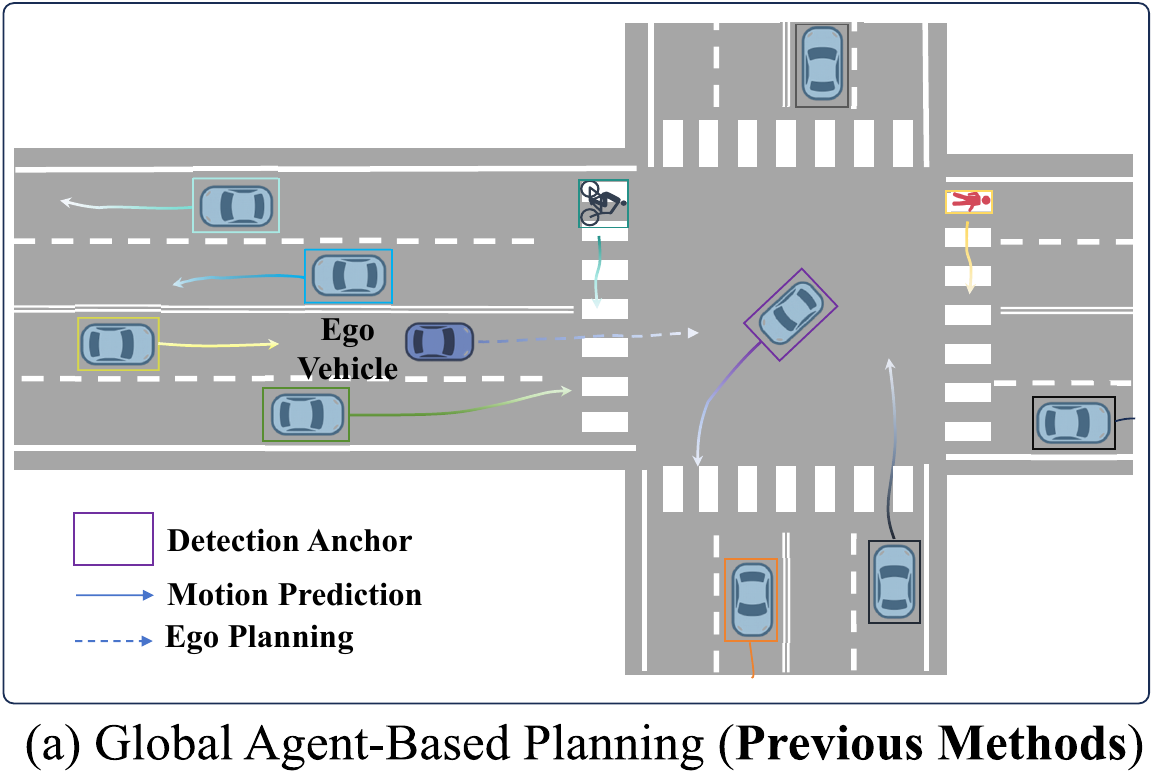}
    \end{subfigure}
    \hfill
    \begin{subfigure}{0.495\linewidth}
        \centering
        \includegraphics[width=\linewidth]{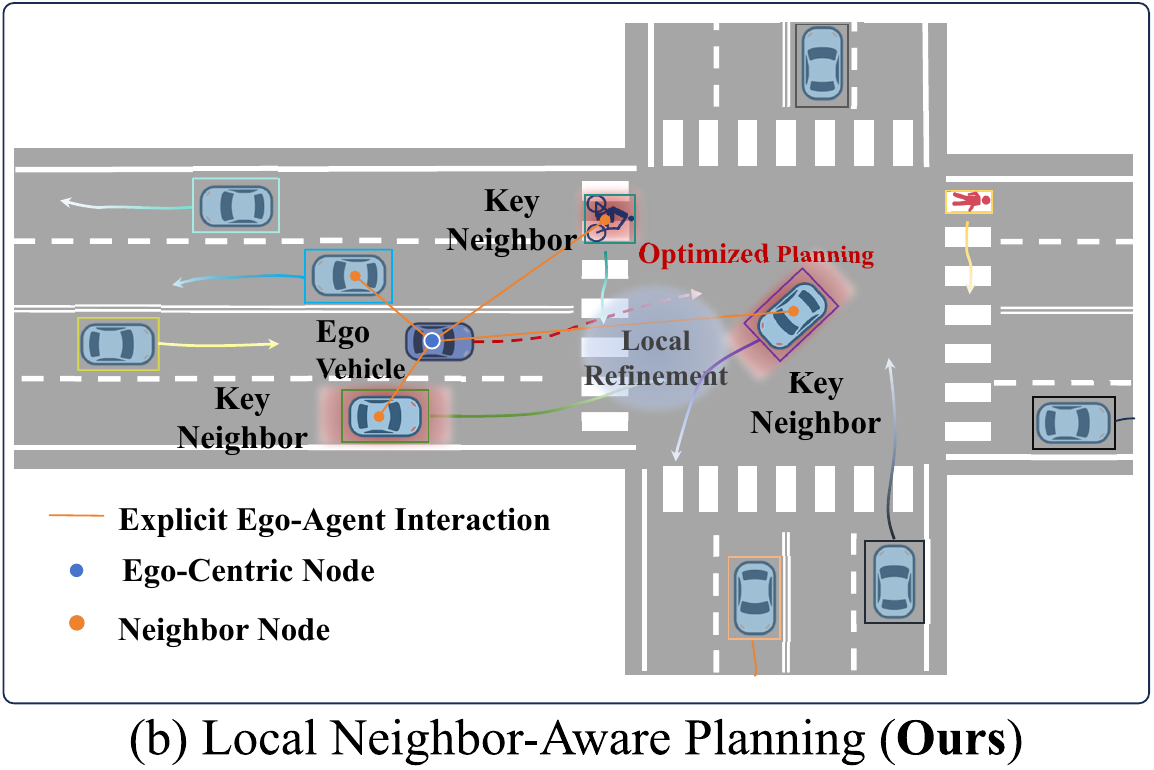}

    \end{subfigure}
    \caption{
        Comparison of global and local planning strategies.
        (a) Global agent-based planning methods~\cite{huPlanningOrientedAutonomousDriving, jiangVADVectorizedScene2023, sunSparseDriveEndtoEndAutonomous2024,liaoDiffusionDriveTruncatedDiffusion2024,xingGoalFlowGoalDrivenFlow2025,chenVADv2EndtoEndVectorized2024,zhangGraphADInteractionScene2024,zhang2025bridging, song2025don} often  overlook critical  local interactions that lead to suboptimal decisions and obscure nearby potential risks.
        (b) We propose FocalAD, which conducts local neighbor-aware planning to highlight critical nearby agents, enabling interaction-aware refinement and safer decision-making.
    }
    \label{fig:motivation}
\end{figure}

End-to-end autonomous driving has emerged as a promising paradigm that directly maps raw sensory inputs to driving actions using fully differentiable unified models~\cite{chenEndtoEndAutonomousDriving2024, chib2023recent}. In traditional modular pipelines, perception~\cite{liBEVFormerLearningBirdsEyeView2022, wang2023exploring, linSparse4DV3Advancing2023}, prediction~\cite{shi2022motion, meinhardt2022trackformer}, and planning~\cite{cheng2024rethinking, fan2018baidu} are handled as separate components, often resulting in the accumulation of errors between modules. In contrast, end-to-end frameworks enable global optimization, thereby improving robustness and interpretability and leading to a more streamlined architecture.

Within this paradigm, ego-trajectory planning remains the central and most challenging task, where the quality of the generated trajectory critically depends on accurately modeling the motion of surrounding traffic agents. This necessity has led to a variety of frameworks that differ in how they structure prediction and planning. Sequential paradigms~\cite{jia2023think, huPlanningOrientedAutonomousDriving, jiangVADVectorizedScene2023, chenVADv2EndtoEndVectorized2024, zhangGraphADInteractionScene2024, yeFusionADMultimodalityFusion2023} first predict the future motion of surrounding agents based on bird's-eye-view (BEV) representations and then pass these predictions to a separate planning module. In contrast, parallel frameworks~\cite{linSparse4DV3Advancing2023, liaoDiffusionDriveTruncatedDiffusion2024, song2025don, fu2023interactionnet} generate planning and prediction trajectories concurrently by leveraging shared perceptual features. This design allows for tighter coupling between motion prediction and ego planning, enabling more coherent decision-making in dynamic environments.  While these approaches have advanced planning performance, they commonly rely on globally aggregated motion features and lack explicit mechanisms to identify the agents that are most critical to ego decision-making, as shown in Fig. \ref{fig:motivation} (a). However, in practical driving scenarios, ego-vehicle planning behavior is primarily influenced by a limited set of nearby agents whose motions have direct and immediate impacts on its decisions. These agents often engage in real-time interactions with the ego vehicle, such as merging, yielding, or crossing, and therefore constitute the main sources of planning risk and constraint. For end-to-end frameworks, the absence of specialized, interaction-aware modeling can lead to the neglect of crucial local cues, impairing the model’s ability to reason effectively in dynamic traffic contexts. As a result, planning output may suffer from reduced interpretability, reliability, and safety, particularly in dense or complex environments.

Drawing inspiration from experienced human drivers, human drivers intuitively prioritize nearby agents over those farther away. Therefore, we argue that end-to-end autonomous driving methods should explicitly be guided to focus on motion cues from critical local agents to refine ego trajectory generation. 

In this work, we propose \textbf{FocalAD}, an end-to-end autonomous driving framework that reinforces local motion awareness, as illustrated in Fig. \ref{fig:motivation} (b). This enhancement is realized through two aspects: enriching planning and motion queries with interaction-aware features and introducing a focal loss to guide attention toward critical neighbor motion features during training. Specifically, FocalAD integrates two tightly coupled modules. \textbf{The Ego-Local-Agents Interactor (ELAI)} constructs representations of ego-centric interaction by explicitly modeling the motion dynamics between the ego vehicle and its local neighbors through a graph-based structure. It focuses on capturing fine-grained, localized interactions that are most relevant to the ego vehicle’s decision-making process. \textbf{The Focal-Local-Agents Loss (FLA Loss)} introduces a focal supervision mechanism that directs the model attention toward decision-critical agents. By leveraging motion cues from influential neighbors, it reinforces high-impact interactions during training, to prioritize agents that significantly affect the ego vehicle's future trajectory. This synergy between representation and supervision improves motion understanding, leading to improved planning safety and interpretability.

Experimental results on nuScenes and Bench2Drive  demonstrate that FocalAD outperforms the baseline~\cite{sunSparseDriveEndtoEndAutonomous2024}. Importantly, on the challenging and complex Adv-nuScenes dataset, FocalAD reduces the collision rate by \textbf{41.9\%} relative to DiffusionDrive~\cite{liaoDiffusionDriveTruncatedDiffusion2024} and by \textbf{15.6\%} relative to SparseDrive~\cite{sunSparseDriveEndtoEndAutonomous2024}, highlighting its superior robustness in challenging driving scenarios. These results validate the effectiveness of our focal-interaction-centric design and underscore the value of interaction-aware learning for safe and robust autonomous planning.


\section{Related Work}\label{sec2}

End-to-end autonomous driving has progressed rapidly, with research priorities gradually shifting from unified perception-control learning toward task decoupling and modular end-to-end design~\cite{chenEndtoEndAutonomousDriving2024}. Transformer-based architectures have significantly improved feature representation by projecting multiview images into the bird's eye view (BEV) space, allowing unified and efficient spatial understanding~\cite{chen2020end, teng2022hierarchical, jia2023driveadapter, liBEVFormerLearningBirdsEyeView2022, li2022panoptic}. ThinkTwice~\cite{jia2023think} highlights the overlooked role of the decoder in tasks such as scene prediction and risk assessment, proposing a cascaded decoder design but still lacks full end-to-end integration. UniAD~\cite{huPlanningOrientedAutonomousDriving} enables joint optimization of perception and planning through a planning-oriented end-to-end framework based on dense BEV representations. This design facilitates the flow of information across stages, thus enhancing the overall performance of planning. Beyond dense BEV-based frameworks, VAD~\cite{jiangVADVectorizedScene2023} introduces a vectorized scene representation that encodes road boundaries, lane markings, and agent trajectories as structured vectors, improving interpretability and controllability while reducing redundancy. SparseDrive~\cite{sunSparseDriveEndtoEndAutonomous2024} introduces a sparse symmetric perception architecture that encodes only key agents and map elements and adopts a parallel motion-planning pipeline to reduce computational overhead without sacrificing performance. In addition to architectural parallelism, PPAD~\cite{chen2024ppad} introduces temporal parallelism by interleaving planning and prediction at each timestep, enabling bidirectional coupling that accounts for evolving agent interactions. To enhance multi-modal trajectory optimization, VADv2~\cite{chenVADv2EndtoEndVectorized2024} introduces a probabilistic planning mechanism that models clustered human trajectories as distributions, improving diversity and predictive capability in future scenarios. DiffusionDrive~\cite{liaoDiffusionDriveTruncatedDiffusion2024} employs an anchored Gaussian prior and two-stage denoising within a diffusion framework to improve trajectory accuracy and controllability in end-to-end driving. GenAD~\cite{zheng2024genad} formulates trajectory prediction and planning as a unified generative modeling task by constructing a structured latent space using variational autoencoders (VAE) and employs temporal GRUs to better model interactive dynamics. In the context of interaction-aware scene understanding, GraphAD~\cite{zhangGraphADInteractionScene2024} proposes a unified graph-based framework that models spatial relationships among the ego vehicle, surrounding agents, and map elements. This design enhances interaction reasoning and improves decision-making quality. Furthermore, FASIONAD++~\cite{qianFASIONADIntegratingHighLevel2025}, inspired by the dual-process theory of cognition (“fast and slow thinking”), introduces a dual system architecture that combines a fast end-to-end planner with a slower logic module based on a vision language model (VLM).

While previous works have substantially improved planning performance, most frameworks still rely on globally aggregated motion features for trajectory generation. However, neglecting localized interactions can obscure critical risks, leading to suboptimal or unsafe plans. To address this, we introduce FocalAD, an end-to-end framework that leverages motion information from key local interactions to refine planning decisions.


\section{Method}\label{sec3}

\begin{figure}
    \centering
    \includegraphics[width=0.98\linewidth]{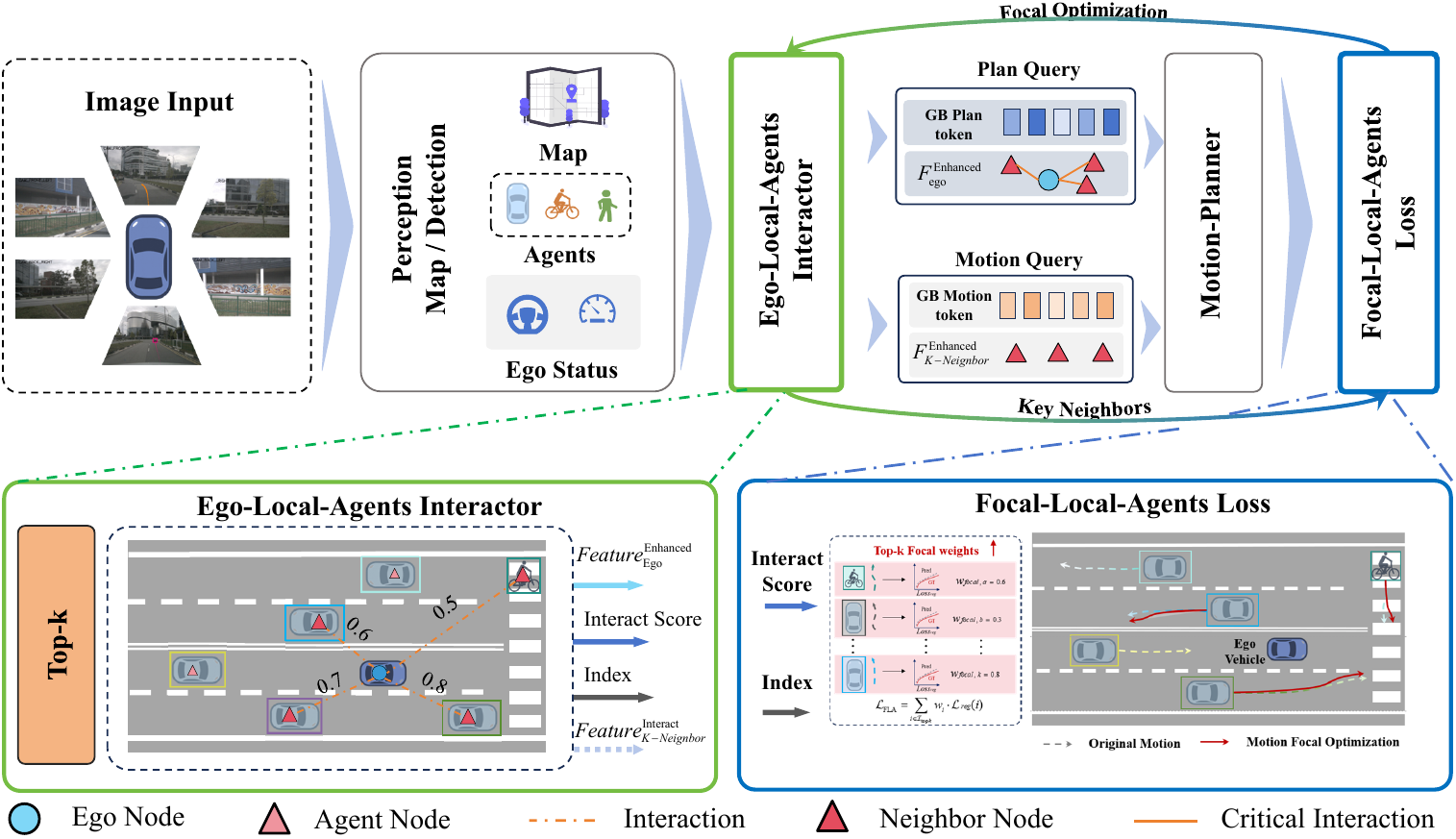}
    \caption{The overall architecture of FocalAD. FocalAD extracts ego and agent states from detection results, enriches planning and motion queries through an Ego-Local-Agents Interactor, and applies a Focal-Local-Agents Loss that assigns focal weights to optimize local motion. This design reinforces local motion awareness through interaction-guided learning, leading to improved planning safety and interpretability.}
    \label{fig:architecture}
\end{figure}

\textbf{Framework Overview.}
Unlike previous methods that rely on globally aggregated motion features, our \textbf{FocalAD} focuses on a limited set of local agents whose behaviors have immediate and significant influence on ego-vehicle planning. To capture and emphasize these localized interactions, FocalAD establishes an interaction-driven mechanism that couples feature representation with loss supervision. As illustrated in \ref{fig:architecture}, FocalAD consists of two core modules: (1) The \textbf{Ego-Local-Agents Interactor (ELAI)} (Sec.~\ref{ELAI}) explicitly models ego-centric interactions through a graph structure. It captures the dynamics between the ego vehicle and its Top-k most relevant neighbors, thereby producing structurally enriched features for both motion and plan queries. (2) The \textbf{Focal-Local-Agents Loss (FLA Loss)} (Sec.~\ref{FLA Loss}) assigns focal weights to key neighboring agents based on their interaction scores and indices, and applies interaction-guided supervision to optimize the learning of neighbor motion features. This forms an interaction-aware mechanism that continuously refines motion and planning representations by aligning model attention with decision-critical motion cues. Together, these components strengthen FocalAD’s ability to reason about local risks and understand dynamic driving contexts, thus improving trajectory prediction precision, planning robustness, and overall interpretability.

\subsection{Ego-Local-Agents Interactor}\label{ELAI}

\begin{figure}
    \centering
    \includegraphics[width=0.98\linewidth]{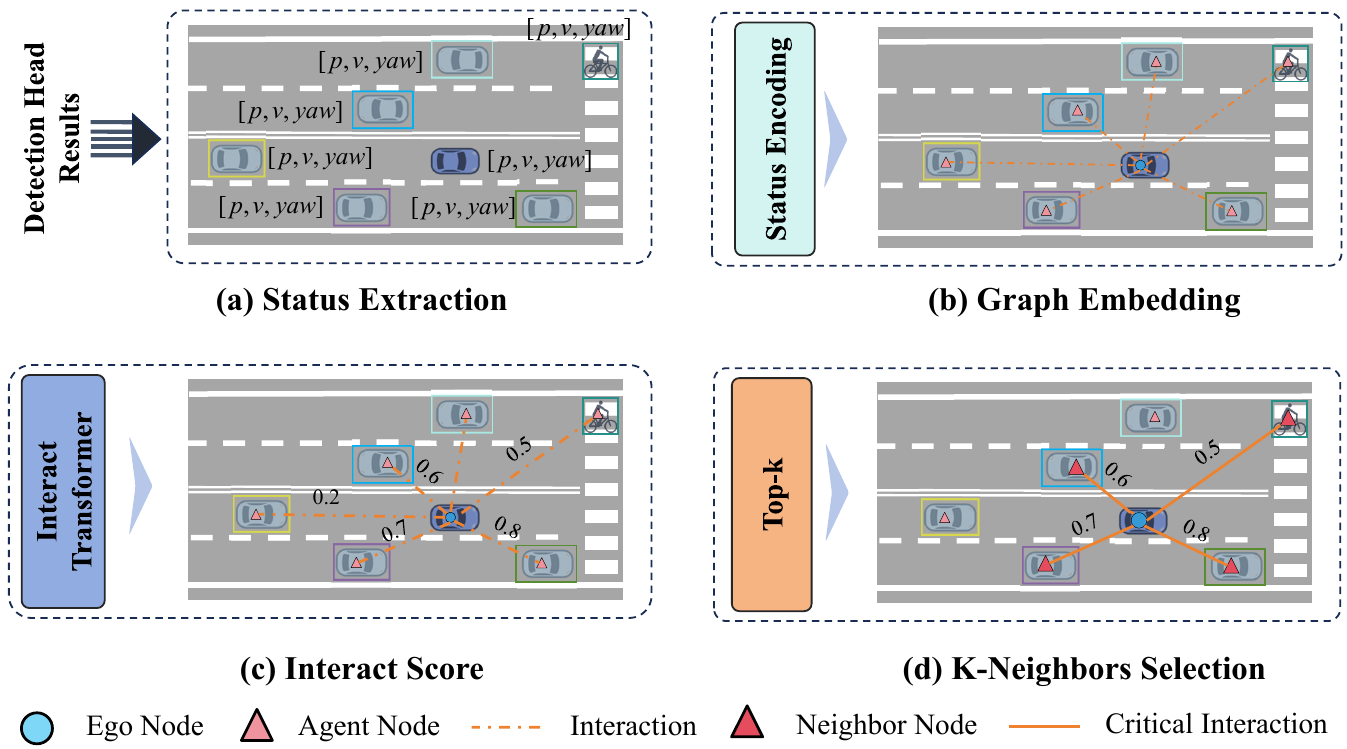}
    \caption{Ego-Local-Agents Interactor. This module enhance ego and agent features by explicit modeling ego-centric interactions. (a) The dynamic states(position, velocity, and heading) of the ego vehicle and surrounding agents are extracted from detection results. (b) These states are encoded into a interaction graph. (c) An Interact Transformer computes pairwise interaction scores. (d) The top-k critical neighbors with the highest interaction scores are selected to enrich motion and plan queries, which are then forwarded for planning and loss supervision.}
    \label{fig:ELAI}
\end{figure}

The Ego-Local-Agents Interactor (ELAI) is designed to explicitly model localized interactions between the ego vehicle and its surrounding agents, enabling interaction-aware ego planning. As shown in Fig.~\ref{fig:ELAI}, ELAI consists of four main steps: Status Extraction, Graph Embedding, Interact Score, and K-Neighbor Selection. 

\noindent\textbf{Status Extraction.} The process begins by extracting the dynamic states of the ego vehicle and agents from detection outputs. Each agent, including the ego, is represented by its kinematic and positional features, which serve as initial inputs for interaction modeling. 

\noindent\textbf{Graph Embedding.} To construct the interaction graph, we first encode the individual motion state of each agent into a node feature representation. Specifically, the node feature for each agent $\mathbf{h}_i$ is computed:

\begin{equation}
\mathbf{h}_{i}=\mathbf{Encoder}_{node}(\left[\mathbf{p}_{i},\mathbf{s}_{i}, \theta_{i}, \mathbf{v}_{i}\right])
\end{equation}
where $\mathbf{p}_i, \mathbf{s}_i, \mathbf{v}_i \in \mathbb{R}^2$ denote the position, size and velocity vectors of agent $i$, respectively, and $\theta_i \in \mathbb{R}$ denotes the heading angle of agent $i$. To model pairwise interactions, we define directed edges from each agent $i$ to the ego vehicle and compute the corresponding edge features:

\begin{equation}
\mathbf{e}_{ego,i} =\mathbf{Encoder}_{edge} ([\Delta\mathbf{p}_{ego,i}, \Delta\theta_{i},\Delta\mathbf{v}_{ego,i}])
\end{equation}
where $\Delta \mathbf{p}_{ego, i}$, $\Delta\theta_{i}$ and $\Delta \mathbf{v}_{ego, i}$ denote the relative position, heading and velocity between agent $i$ with respect to the ego vehicle.

\noindent\textbf{Interact Score.} An Interact Transformer is employed to compute pairwise interaction scores between the ego vehicle and surrounding agents, capturing their potential influence on ego decision-making. We first apply a multi-head attention mechanism to aggregate local interaction features, where the ego vehicle feature $\mathbf{h}_{ego}$ is treated as the query vector, and each agent combined node edge feature forms the key and value vectors:
\begin{equation}
\mathbf{c}_{ego}=\mathrm{MHCA}(\mathbf{h}_{\mathrm{ego}}, [\mathbf{h}_{i}\parallel\mathbf{e}_{\mathrm{ego},i}],[\mathbf{h}_{i}\parallel\mathbf{e}_{\mathrm{ego},i}] ).
\end{equation}
where MHCA denotes multi-head cross-attention and $\mathbf{c}_{ego}$ denotes the ego-centric interaction context. This context encodes planning-relevant motion semantics by capturing relative dynamics, spatial relationships, and interaction intensities between the ego vehicle and agents. To quantify each agent’s impact on planning, a multilayer perceptron (MLP) processes the of its raw motion feature $\mathbf{h}_i$, the edge feature $\mathbf{e}_{ego,i}$, and the shared interaction context $\mathbf{c}_{ego}$. It jointly outputs an enhanced feature representation $\mathbf{h}_i^{\mathrm{enc}}$ and its interaction score $s_i$ indicating the agent’s contribution to ego decision-making:
\begin{equation} 
\left(\mathbf{h}_i^{\mathrm{enc}}, s_i\right) = \mathrm{MLP}\left([\mathbf{h}_i \parallel \mathbf{e}_{\mathrm{ego},i}   \parallel  \mathbf{c}_{\mathrm{ego}} ]\right). 
\end{equation}

\noindent\textbf{K-Neighbors Selection.}
The Top-k most relevant neighbors, identified via interaction scores, are indexed by the set $\mathcal{I}_{\mathrm{top-}k}$. The associated cirtical neighbor features are represented $\mathbf{h}_i^{N}$, which represents the interaction-aware motion feature for critical neighbor $i$. To incorporate local interaction feature, the global motion query $\mathbf{Q}_{motion,i}$ for each agent is refined using interaction-aware modifications from significant neighbors as follows:
\begin{equation}
    \mathbf{Q}_{motion,i}^{\mathrm{ref}}=
\begin{cases}
 \mathbf{Q}_{motion,i}+\gamma\cdot\alpha_i\cdot\mathbf{h}_i^{N}, & \mathrm{if} i\in\mathcal{I}_{\mathrm{top-}k} \\
 \mathbf{Q}_{motion,i}, & \mathrm{otherwise} 
\end{cases}
\end{equation}
where $\gamma$ is a scaling factor and $\alpha_i$ is the attention weight obtained by applying softmax to the score $s_i$. 
To strengthen the representation of local interactions in the planning query, the ego representation 
$\mathbf{h}_{\text{ego}}$ is refined by fusing it with the critical neighbor feature $\mathbf{h}_i^{N}$.
\begin{equation}
    \mathbf{h}_{ego}^{\prime}=\mathrm{MLP}\left(\left[\mathbf{h}_{ego}\parallel\mathbf{h}_i^{N}\right]\right).
\end{equation}
This enhanced ego representation captures structured interaction semantics in the local scene and serves as the foundation for downstream computation.
The global planning query $\mathbf{Q}_{plan}$ is refined using the updated $\mathbf{h}_{\mathrm{ego}}^{\prime}$:
\begin{equation}
    \mathbf{Q}_{plan}^{\mathrm{ref}}=\mathbf{Q_{plan}}+\beta\cdot\mathbf{h}_{ego}^{\prime}
\end{equation}
where $\beta$ is scaling factors.

\subsection{Focal-Local-Agents Loss}\label{FLA Loss}

\begin{figure}
    \centering
    \includegraphics[width=0.98\linewidth]{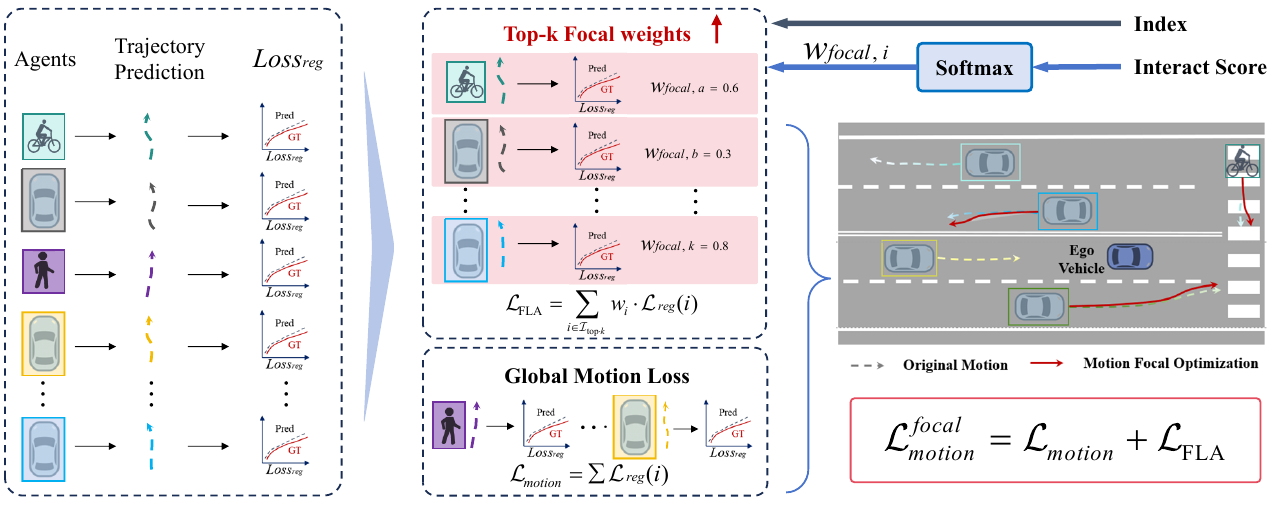}
    \caption{Focal-Local-Agents Loss. This module assigns focal weights to critical agents according to their interaction scores, and integrates these weights with a global motion loss to provide decision-aware supervision for motion prediction.}
    \label{fig:FLA}
\end{figure}

In the preceding module, we obtain informative representations for the Top-k critical neighbors. As illustrated in Fig.~\ref{fig:FLA}, the proposed Focal-Local-Agents Loss (FLA Loss) introduces an interaction-aware training strategy that links motion supervision to neighbor feature learning, guiding the model to focus on decision-relevant agents. 

To derive supervision weights for key neighbors, we apply a softmax operation to the interaction scores $s_i$
obtained from the preceding module. This yields normalized focal weights $w^{focal}_i$ for each neighbor $i\in\mathcal{I}_{\mathrm{top}-k}$. These focal weights reflect the estimated influence of the neighbor on ego decision making. Each neighbor $i\in\mathcal{I}_{\mathrm{top}-k}$ corresponds to a trajectory regression loss $\mathcal{L}_{\mathrm{motion}}(i)$, which is sampled according to its index and subsequently weighted by the focal coefficient $w^{focal}_i$. The overall FLA Loss is defined as:
\begin{equation}
\mathcal{L}_{FLA}=\sum_{i\in\mathcal{I}_{\mathrm{top}\cdot k}}w_i\cdot\mathcal{L}_{motion}(i)
\end{equation}
This formulation encourages the model to focus on decision-critical neighbors during training, enabling more targeted motion supervision and interaction-aware feature refinement. To integrate focal guidance into the overall training objective, the final motion loss is defined as a combination of the standard global motion loss and the FLA Loss:
\begin{equation}
    \mathcal{L}_{motion}^{\mathrm{focal}}=\mathcal{L}_{motion}+\mathcal{L}_{FLA}
\end{equation}

Unlike standard supervision strategies that apply uniform weighting, FLA Loss leverages weighting based on planning relevance to enhance both planning reliability and training efficiency in complex interaction settings. This interaction-aware supervision addresses a key limitation of conventional paradigms, which often lack targeted guidance for decision-critical agents. As a result, the model couples key neighbor motions with planning-oriented feature refinement to better capture critical interactions.


\section{Experiments}

\subsection{Datasets and Evaluation Metrics.}
\noindent\textbf{nuScenes.} For open-loop evaluation, we conduct extensive experiments on the  \textbf{nuScenes} dataset~\cite{caesar2020nuscenes}, which contains 1,000 20-second driving scenes with annotations at 2 Hz. It provides rich multimodal sensor data, including six camera views per keyframe, 3D object detection labels, and high-definition semantic maps. 

\noindent\textbf{Bench2Drive.} Closed-loop evaluation is conducted on the \textbf{Bench2Drive}~\cite{jia2024bench2drive}, a large-scale benchmark built on the CARLA simulator ~\cite{dosovitskiy2017carla} under the CARLA Leaderboard 2.0 protocol. It contains more than 2 million frames that span 44 interaction scenarios and 23 weather conditions, allowing realistic and fine-grained evaluation of end-to-end planning in complex environments. We use the official 220 routes for evaluation.

\noindent \textbf{Adv-nuSc.} Beyond regular open-loop and closed-loop evaluations, we further assess robustness on the Adv-nuSc dataset~\cite{xu2025challenger}, which is constructed using the Challenger framework. This extended dataset is built upon nuScenes and is specifically designed to reveal planning vulnerabilities under complex traffic conditions. The Challenger framework generates a diverse set of aggressive driving scenarios, such as sudden cut-ins, abrupt lane changes, tailgating, and blind spot intrusions, and renders them into photorealistic multiview videos with 3D annotations. Adv-nuSc comprises 156 safety-critical scenarios (6,115 samples), each crafted to simulate high-risk interactions that stress-test the decision-making capability of autonomous driving systems.

\noindent\textbf{Planning Evaluation Metrics.} For the planning evaluation, we adopt two commonly used metrics: L2 Displacement Error (L2) and Collision Rate, both computed following the protocol defined in SparseDrive~\cite{sunSparseDriveEndtoEndAutonomous2024} to ensure consistency with previous work.

\subsection{Implementation Details.} To capture dynamic and static instances around the ego vehicle, we adopt a sparse perception module based on SparseDrive~\cite{sunSparseDriveEndtoEndAutonomous2024}. Multiview images are processed by a ResNet-50~\cite{he2016deep} backbone and the input image size is 256×704. The extracted features are aggregated into instance-level representations for traffic agents and map elements at each time step. For interaction modeling, the number of Top-k selected neighbors is set to 5 by default unless otherwise specified. For robustness evaluation, our model is trained on the standard nuScenes training set following common practice. We then separately evaluated it on the adversarial Adv-nuSc dataset to assess its performance under challenging conditions.

\subsection{Main Results}

\begin{table}[t]
\centering
\begin{subtable}[t]{0.455\textwidth}
\centering
\begin{adjustbox}{max width=\linewidth}
\begin{tabular}{l|c|c|c|c}
\toprule
\textbf{Method} & minADE$\downarrow$ & minFDE$\downarrow$ & MR$\downarrow$ & EPA$\uparrow$ \\
\midrule
Traditional~\cite{gu2023vip3d} & 2.06 & 3.02 & 0.277 & 0.209 \\
PnPNet~\cite{liang2020pnpnet} & 1.15 & 1.95 & 0.226 & 0.222 \\
ViP3D~\cite{gu2023vip3d} & 2.05 & 2.84 & 0.246 & 0.226 \\
UniAD~\cite{huPlanningOrientedAutonomousDriving} & 0.71 & 1.02 & 0.151 & 0.456 \\
SparseDrive~\cite{sunSparseDriveEndtoEndAutonomous2024} & 0.62 & 0.99 & 0.136 & 0.482 \\

\rowcolor[rgb]{0.902,0.902,0.902}\textbf{FocalAD (Ours)} & \textbf{0.61} & \textbf{0.95} & \textbf{0.134} & \textbf{0.490} \\
\bottomrule
\end{tabular}
\end{adjustbox}
\caption{Motion prediction results.}
\end{subtable}
\hfill
\begin{subtable}[t]{0.525\textwidth}
\centering
\begin{adjustbox}{max width=\linewidth}
\begin{tabular}{l|cccc|cccc}
\toprule
\textbf{Method} & \multicolumn{4}{c|}{L2($m$)$\downarrow$} & \multicolumn{4}{c}{Col. Rate(\%)$\downarrow$} \\
 & 1s & 2s & 3s & Avg & 1s & 2s & 3s & Avg \\
\midrule
ST-P3~\cite{hu2022st} & 1.33 & 2.11 & 2.90 & 2.11 & 0.23 & 0.62 & 1.27 & 0.71 \\
UniAD~\cite{huPlanningOrientedAutonomousDriving} & 0.48 & 0.96 & 1.65 & 1.03 & 0.10 & 0.15 & 0.61 & 0.29 \\
VAD$^\dagger$~\cite{jiangVADVectorizedScene2023} & 0.41 & 0.70 & 1.05 & 0.72 & 0.11 & 0.24 & 0.42 & 0.26 \\
SparseDrive$^\dagger$~\cite{sunSparseDriveEndtoEndAutonomous2024} & 0.30 & 0.58 & 0.96 & 0.61 & 0.01 & 0.05 & 0.23 & 0.10 \\
DiffusionDrive$^\dagger$~\cite{liaoDiffusionDriveTruncatedDiffusion2024} & 0.29 & 0.58 & 0.96 & 0.61 & 0.02 & 0.05 & 0.22 & 0.09 \\

\rowcolor[rgb]{0.902,0.902,0.902}\textbf{FocalAD (Ours)} & \textbf{0.27} & \textbf{0.57} & \textbf{0.96} & \textbf{0.60} & \textbf{0.00} & \textbf{0.04} & \textbf{0.24} & \textbf{0.09} \\
\bottomrule
\end{tabular}
\end{adjustbox}
\caption{Planning results.}
\end{subtable}
\vspace{3pt}
\caption{Motion prediction and planning results on the nuScenes validation set. $\dagger$: Reproduced with official checkpoint.}
\label{tab:motion-planning}
\vspace{-20pt}
\end{table}

\noindent \textbf{nuScenes Results.} As shown in Table~\ref{tab:motion-planning} (a), FocalAD achieves the best overall motion prediction performance across all metrics, reducing minADE, minFDE and MR to \textbf{0.61m} , \textbf{0.95m} and \textbf{0.134}, respectively. In addition, FocalAD achieves the highest EPA of \textbf{0.494}, outperforming all strong baselines. These results highlight the effectiveness of our localized interaction modeling, which enables more accurate and confident multi-agent trajectory predictions. Table~\ref{tab:motion-planning} (b) further shows that FocalAD achieves the lowest average planning error of \textbf{0.60m}, along with an exceptionally low average collision rate of \textbf{0.09\%}, demonstrating performance comparable to the best baselines. Notably, FocalAD achieves superior early-stage planning performance, attaining the lowest L2 errors at both 1s and 2s, and eliminating collisions entirely at 1s. Its 3s performance, while slightly surpassed by SparseDrive and DiffusionDrive, remains competitive. These results demonstrates that FocalAD significantly improves planning performance and safety in local interaction scenarios, particularly in short-horizon decision-making tasks.

\noindent \textbf{Bench2Drive Results.} On the Multi-Ability Benchmark (Table~\ref{tab:closed loop ability}), FocalAD outperforms UniAD-Base, VAD, and SparseDrive across all five critical driving tasks, achieving the highest mean ability score of \textbf{20.53\%}. It consistently ranks first in Merging, Overtaking, Give Way, and Traffic Sign ability, and second in Emergency Brake. On the metric-based evaluation (Table~\ref{tab:closed loop metrics}), FocalAD achieves the best overall performance, with the lowest open-loop L2 error (\textbf{0.85}), the highest Driving Score (\textbf{45.77}), Success Rate (\textbf{17.30\%}), and Efficiency (\textbf{174.01}). Compared to SparseDrive, it improves Driving Score by \textbf{+1.23}, SR by \textbf{+0.59\%}, and reduces the L2 error by \textbf{0.02m}. These results highlight the importance of modeling local interactions for enhancing planning performance.

\begin{table*}[t]
\centering
\begin{minipage}[t]{0.55\linewidth}
\centering
\begin{adjustbox}{max width=\linewidth}
\begin{tabular}{l|c|c|c|c|c|c} 
\toprule
\multirow{2}{*}{\textbf{Method}} & \multicolumn{6}{c}{\textbf{Ability} (\%) $\uparrow$}   \\ 
\cmidrule{2-7}
& Merging        & \makecell{Over\\-taking}     & \makecell{Emergency\\Brake} & \makecell{Give\\Way}       & \makecell{Traffic\\Sign}   & \textbf{Mean}   \\ 
\hline
UniAD-Base~\cite{huPlanningOrientedAutonomousDriving}  & 14.10          & 17.78          & \textbf{21.67}           & 10.00          & 14.21          & 15.55           \\
VAD~\cite{jiangVADVectorizedScene2023}    & 8.11           & 24.44          & 18.64           & 20.00          & 19.15          & 18.07           \\
SparseDrive$^\dagger$~\cite{sunSparseDriveEndtoEndAutonomous2024}  & 12.18           & 23.19          & 17.91           & 20.00          & 20.98          & 17.45           \\  
\rowcolor[rgb]{0.902,0.902,0.902} \textbf{FocalAD (Ours)}  & \textbf{14.11} & \textbf{24.81}  & 20.31 & \textbf{20.00} & \textbf{23.42} & \textbf{20.53} \\
\bottomrule
\end{tabular}
\end{adjustbox}
\caption{Bench2Drive: Multi-Ability Benchmark. $\dagger$: Reproduced with official checkpoint.}
\label{tab:closed loop ability}
\end{minipage}
\hfill
\begin{minipage}[t]{0.43\linewidth}
\centering
\begin{adjustbox}{max width=\linewidth}
\begin{tabular}{l|c|ccc}
\toprule
\textbf{Method} & \textbf{Avg. L2} & \textbf{DS} & \textbf{SR(\%)} & \textbf{Effi}  \\
               & (Open) $\downarrow$ & $\uparrow$ & $\uparrow$ & $\uparrow$ \\
\midrule
VAD~\cite{jiangVADVectorizedScene2023} & 0.91 & 42.35 & 15.00 & 157.94  \\
VADV2~\cite{chenVADv2EndtoEndVectorized2024} & 0.89 & 42.87 & 15.91 & 158.12  \\
SparseDrive$^*$~\cite{sunSparseDriveEndtoEndAutonomous2024} & 0.87 & 44.54 & 16.71 & 170.21  \\
\rowcolor[rgb]{0.902,0.902,0.902} \textbf{FocalAD (Ours)} & \textbf{0.85} & \textbf{45.77} & \textbf{17.30} & \textbf{174.01} \\
\bottomrule
\end{tabular}
\end{adjustbox}
\caption{Bench2Drive: Metric Comparison. $*$: Results reported in ~\cite{song2025don}.} 
\label{tab:closed loop metrics}
\end{minipage}
\vspace{-3mm}
\end{table*}

\begin{table*}[t]
\centering
\begin{minipage}[t]{0.58\linewidth}
\centering
\begin{adjustbox}{max width=\linewidth}
\begin{tabular}{l|c|ccc}
\toprule
\textbf{Method} & minADE$\downarrow$ & minFDE$\downarrow$ & MR$\downarrow$ & EPA$\uparrow$ \\
\midrule
SparseDrive$^\dagger$~\cite{sunSparseDriveEndtoEndAutonomous2024} & 1.12 & 1.62 & 0.25 & 0.15  \\
DiffusionDrive$^\dagger$~\cite{liaoDiffusionDriveTruncatedDiffusion2024} & 1.23 & 1.83 & 0.27 & 0.14  \\
\rowcolor[rgb]{0.902,0.902,0.902} \textbf{FocalAD (Ours)} & \textbf{1.07} & \textbf{1.58} & \textbf{0.18} & \textbf{0.16} \\
\bottomrule
\end{tabular}
\end{adjustbox}
\caption{Motion prediction  on Adv-nuSc$\dagger$: Reproduced with official checkpoint.}
\label{tab:motion on adv}
\end{minipage}
\hfill
\begin{minipage}[t]{0.4\linewidth}
\centering
\begin{adjustbox}{max width=\linewidth}
\begin{tabular}{l|cccc}
\toprule
\textbf{AD Model} & 1s & 2s & 3s & avg. \\
\midrule
UniAD~ & 0.80\% & 4.10\% & 6.96\% & 3.95\% \\
VAD~ & 4.46\% & 7.59\% & 9.08\% & 7.05\% \\
SparseDrive$^\dagger$~\cite{sunSparseDriveEndtoEndAutonomous2024} & 0.06\% & 0.95\% & 2.44\% & 1.15\% \\
DiffusionDrive$^\dagger$~\cite{liaoDiffusionDriveTruncatedDiffusion2024} & 0.08\% & 1.23\% & 3.65\% & 1.67\% \\
\rowcolor[rgb]{0.902,0.902,0.902} \textbf{FocalAD (Ours)} & \textbf{0.05\%} & \textbf{0.57\%} & \textbf{2.28\%} & \textbf{0.97\%} \\
\bottomrule
\end{tabular}
\end{adjustbox}
\caption{Collision rates on Adv-nuSc$\dagger$: Reproduced with official checkpoint.}
\label{tab:robustness}
\end{minipage}
\vspace{-3mm}
\end{table*}

\subsection{Robustness Analysis}

\begin{wrapfigure}{r}{0.48\linewidth}
  \centering
  \vspace{-8pt}  
  \includegraphics[width=\linewidth]{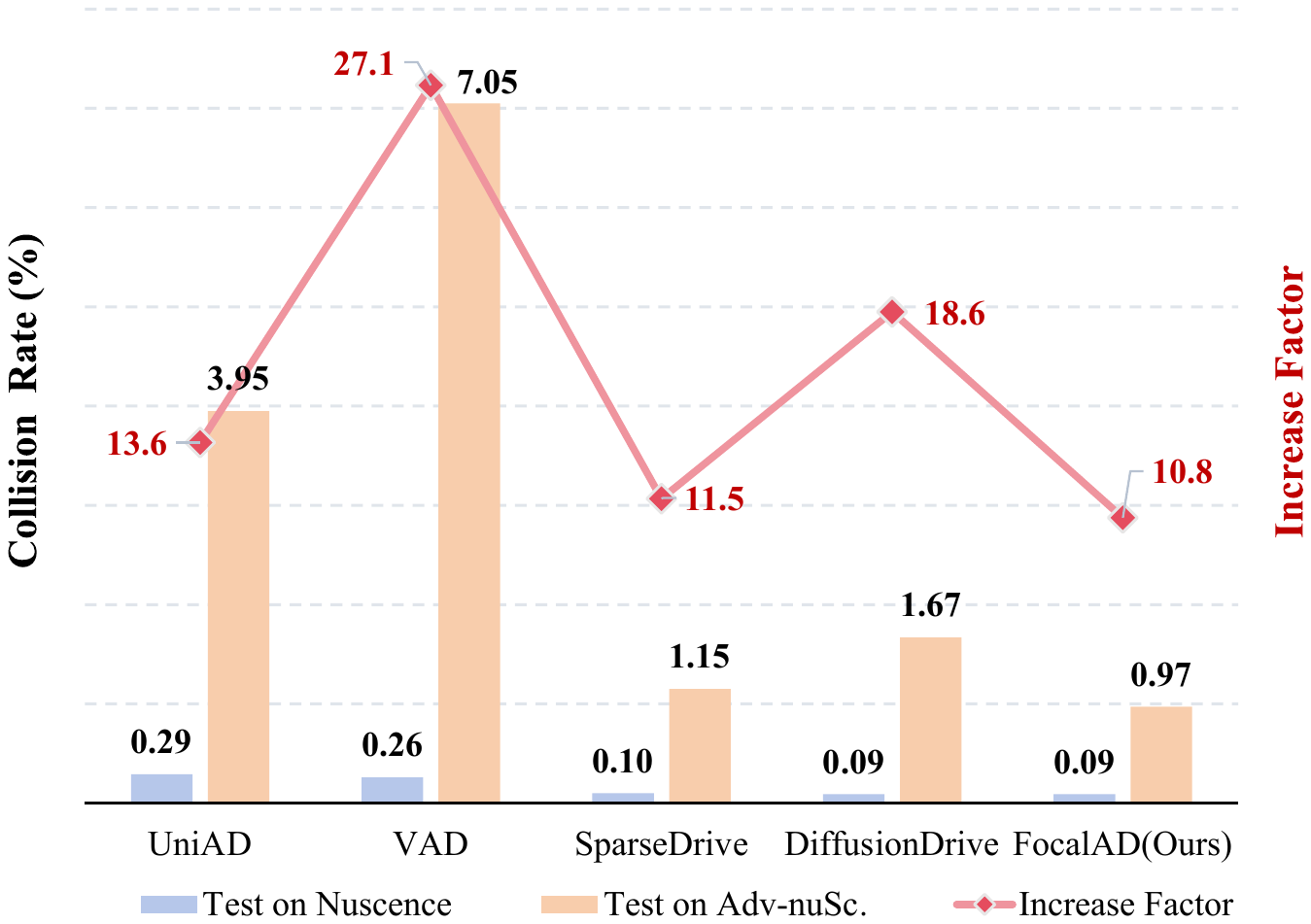}
  \caption{Comparison of collision rates and increase factors on nuScenes and Adv-nuSc. The beige bars represent the collision rates (\%) on each dataset, while the red line denotes the increase factor, calculated as the ratio of the collision rate on Adv-nuSc to that on nuScenes.}
  \label{fig:IncreaseRate}
  \vspace{-20pt}  
\end{wrapfigure}

Although large-scale datasets such as nuScenes~\cite{caesar2020nuscenes},  NAVSIM~\cite{dauner2024navsim}, and Bench2Drive~\cite{jia2024bench2drive} have enabled substantial advances in the evaluation of autonomous driving systems in various scenarios, they consist predominantly of natural traffic flows. As a result, they lack coverage of rare but critically critical interactions that pose significant challenges to planning and decision-making. This limitation hinders a systematic assessment of model robustness under under high-risk or complex driving scenarios. To address this gap, we conduct additional evaluation on the Adv-nuSc dataset~\cite{xu2025challenger}. We compare model performance on both the original nuScenes validation set and the Adv-nuSc dataset to assess how well our method maintains planning reliability when faced with aggressive, unexpected, or highly interactive driving behaviors.

Table~\ref{tab:motion on adv} compares the motion prediction performance of different methods on the Adv-nuSc$\dagger$ dataset based on officially released checkpoints. \textbf{FocalAD} outperforms both SparseDrive and DiffusionDrive across all evaluation metrics, demonstrating superior accuracy, lower miss rates, and improved endpoint alignment. The experimental results in Table~\ref{tab:robustness} demonstrate that FocalAD exhibits superior planning performance in complex or challenging traffic scenarios, achieving a \textbf{41.9\%} reduction in collision rate compared to DiffusionDrive. Specifically, the average collision rate of \textbf{FocalAD} increases from \textbf{0.09\%} on nuScenes to \textbf{0.97\%} on Adv-nuSc, which is the lowest among all evaluated methods. In contrast, state-of-the-art baselines such as SparseDrive and DiffusionDrive suffer from more significant degradations, with average collision rates rising from \textbf{0.10\%} to \textbf{1.03\%} and \textbf{0.09\%} to \textbf{1.67\%}, respectively. VAD experiences the most dramatic increase, from \textbf{0.26\%} to \textbf{7.05\%}, while UniAD rises from \textbf{0.63\%} to \textbf{3.95\%}. A more intuitive comparison of these increases is illustrated in Figure~\ref{fig:IncreaseRate}. Notably, \textbf{FocalAD} exhibits the smallest increase factor of \textbf{10.8$\times$} in collision rate from nuScenes to Adv-nuSc, further demonstrating its robustness under challenging traffic scenarios. 

These results confirm that FocalAD not only maintains strong planning performance in standard conditions but also generalizes significantly better under safety-critical scenarios, highlighting its superior robustness in dynamic urban environments.

\subsection{Ablation Study}
\begin{table}[t]
\centering
\setlength{\extrarowheight}{0pt}
\addtolength{\extrarowheight}{\aboverulesep}
\addtolength{\extrarowheight}{\belowrulesep}
\setlength{\aboverulesep}{0pt}
\setlength{\belowrulesep}{0pt}

\begin{tabularx}{\textwidth}{cc|c|XXXX|XXXX|XXXX}
\toprule
\multirow{3}{*}{\textbf{ELAI}} &
\multirow{3}{*}{\shortstack{\textbf{FLA} \\ \textbf{Loss}}}&
\multirow{3}{*}{\textbf{Top-k}} &
\multicolumn{8}{c|}{\textbf{nuScenes}} &
\multicolumn{4}{c}{\textbf{Adv-nuSc}} \\
\cmidrule(lr){4-11} \cmidrule(lr){12-15}
& & &
\multicolumn{4}{c|}{\textbf{L2 (m)} $\downarrow$} &
\multicolumn{4}{c|}{\textbf{Collision (\%)} $\downarrow$} &
\multicolumn{4}{c}{\textbf{Collision (\%)} $\downarrow$} \\
& & &
1s & 2s & 3s & \cellcolor[rgb]{0.851,0.851,0.851}Avg. &
1s & 2s & 3s & \cellcolor[rgb]{0.851,0.851,0.851}Avg. &
1s & 2s & 3s & \cellcolor[rgb]{0.851,0.851,0.851}Avg. \\
\midrule
--          & --       & --   & 0.30 & 0.60 & 0.97 & \cellcolor[rgb]{0.851,0.851,0.851}0.63 
                              & 0.02 & 0.12 & 0.36 & \cellcolor[rgb]{0.851,0.851,0.851}0.16 
                              & 0.10 & 0.76 & 2.67 & \cellcolor[rgb]{0.851,0.851,0.851}1.17 \\
\checkmark   & --       & 3   & 0.31 & 0.59 & 0.96 & \cellcolor[rgb]{0.851,0.851,0.851}0.62 
                              & 0.00 & 0.08 & 0.33 & \cellcolor[rgb]{0.851,0.851,0.851}0.14 
                              & 0.05 & 0.72 & 2.59 & \cellcolor[rgb]{0.851,0.851,0.851}1.12 \\
\checkmark   & --       & 5   & 0.31 & 0.59 & 0.95  & \cellcolor[rgb]{0.851,0.851,0.851}0.62 
                              & 0.00 & 0.07 & 0.32 & \cellcolor[rgb]{0.851,0.851,0.851}0.12 
                              & 0.04 & 0.72 & 2.57 & \cellcolor[rgb]{0.851,0.851,0.851}1.11 \\
\checkmark   & --       & 7   & 0.31 & 0.60 & 0.95 & \cellcolor[rgb]{0.851,0.851,0.851}0.63 
                              & 0.01 & 0.06 & 0.32 & \cellcolor[rgb]{0.851,0.851,0.851}0.12 
                              & 0.06 & 0.73 & 2.60 & \cellcolor[rgb]{0.851,0.851,0.851}1.13 \\
\checkmark  & \checkmark & 3  & 0.30 & 0.59 & 0.95 & \cellcolor[rgb]{0.851,0.851,0.851}0.62 
                              & 0.00 & 0.05 & 0.27 & \cellcolor[rgb]{0.851,0.851,0.851}0.11 
                              & 0.06 & 0.58 & 2.34 & \cellcolor[rgb]{0.851,0.851,0.851}1.00  \\
\midrule
\checkmark  & \checkmark & 5  & 0.27 & 0.57 & 0.96 & \cellcolor[rgb]{0.851,0.851,0.851}0.60 
                              & 0.00 & 0.04 & 0.24 & \cellcolor[rgb]{0.851,0.851,0.851}0.09 
                              & 0.05 & 0.57 & 2.29 & \cellcolor[rgb]{0.851,0.851,0.851}0.97  \\
\midrule
\checkmark & \checkmark  & 7  & 0.30 & 0.60 & 0.98 & \cellcolor[rgb]{0.851,0.851,0.851}0.63
                              & 0.00 & 0.05 & 0.28 & \cellcolor[rgb]{0.851,0.851,0.851}0.11 
                              & 0.05 & 0.60 & 2.33 & \cellcolor[rgb]{0.851,0.851,0.851}0.99 \\
\bottomrule
\end{tabularx}
\caption{Ablation studies of the impact of the different modules on the nuScenes validation dataset. We follow the SparseDrive~\cite{sunSparseDriveEndtoEndAutonomous2024} evaluation metric.}
\label{tab:representation-ablation}
\end{table}


To evaluate the contribution of each proposed component, we conduct ablation studies on the nuScenes and Adv-nuSc validation sets, as shown in Table~\ref{tab:representation-ablation}. 

\noindent\textbf{nuScenes validation set.} We first evaluate the impact of the \textbf{ELAI} module. When enabling ELAI alone (without FLA Loss), the model consistently improves both planning accuracy and safety over the baseline, reducing the average L2 error to \textbf{0.62m} and the average collision rate to \textbf{0.12\%}. This result highlights the value of explicitly modeling ego-centric interactions with local agents. Introducing the \textbf{FLA Loss} yields further improvements, particularly in safety. With \texttt{Top-k = 5}, the collision rate drops to \textbf{0.09\%}, while the average L2 error decreases to \textbf{0.60m}. This demonstrates that FLA Loss effectively guides the model to focus on decision-critical neighbors, enhancing motion and planning representations through interaction-aware supervision. To further validate this configuration, we experiment with different \texttt{Top-k} values and find that \texttt{Top-k = 5} consistently yields the best overall performance. 

\noindent\textbf{Adv-nuSc validation set.} A similar trend is observed on the \textbf{Adv-nuSc} dataset, which involves more complex and risk-prone scenarios. The combination of ELAI and FLA Loss with \texttt{Top-k = 5} achieves the lowest average collision rate (\textbf{0.97\%}), improved from \textbf{1.17\%}, among all ablated variants. The larger performance gaps on Adv-nuSc further underscore the stronger contribution of each component under challenging conditions, offering more compelling evidence of their effectiveness and confirming the robustness and generalizability of our approach.

\subsection{Qualitative Analysis}

\begin{figure}[htbp]
    \centering
    \includegraphics[width=1\linewidth]{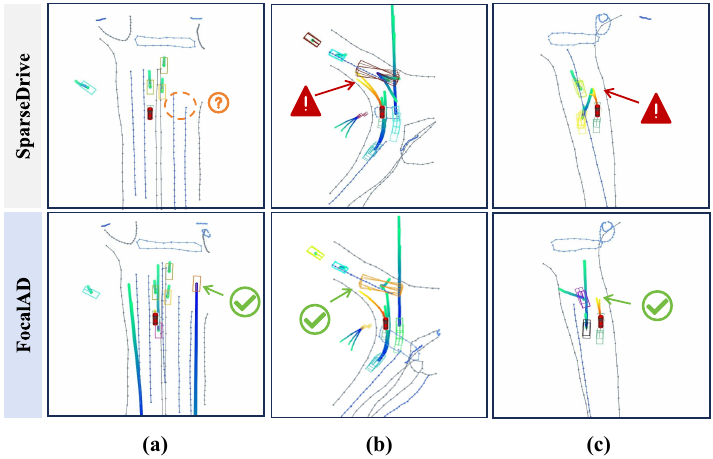}
    \caption{Qualitative Comparison of SparseDrive and FocalAD in Complex Traffic Scenarios. (a) SparseDrive fails to capture local agent interactions, whereas FocalAD effectively establishes interaction links with nearby agents.  (b) SparseDrive fails to yield to a turning bus, while FocalAD safely adjusts its plan. (c) SparseDrive fails to respond to the sudden cut-in, risking a collision. FocalAD accurately anticipates the maneuver and plans a safe trajectory.}
    \label{fig:qualitative}
\end{figure}

To better understand the behavioral differences between SparseDrive and our proposed FocalAD model, we conduct a qualitative analysis on representative scenarios from the Adv-nuScenes dataset. In Fig.~\ref{fig:qualitative} (a), SparseDrive fails to capture local agent interactions, overlooking potential risks in the environment. In contrast, FocalAD accurately models nearby agents, providing planning with a broader sense of risk awareness.
In Fig.~\ref{fig:qualitative} (b), a nearby bus initiates a right turn at an intersection. SparseDrive fails to yield, generating a conflicting trajectory, while FocalAD successfully anticipates the bus intent and adjusts its plan accordingly. In Fig.~\ref{fig:qualitative} (c), a sudden cut-in by a neighboring vehicle is depicted. SparseDrive maintains its original path, risking a collision, whereas FocalAD promptly recognizes the maneuver and adapts its trajectory to ensure safety. These comparisons demonstrate that FocalAD improves planning safety and interpretability by explicitly modeling local agent interactions.

\section{Conclusion}\label{sec13}

In this paper, we presented \textbf{FocalAD}, an end-to-end autonomous driving framework that enhances planning by explicitly modeling critical local motion interactions. Unlike prior approaches that rely on globally aggregated features, FocalAD leverages ego-centric interaction representation through the \textbf{Ego-Local-Agents Interactor (ELAI)} and introduces a interaction-aware training mechanisms via the \textbf{Focal-Local-Agents Loss (FLA Loss)}. Extensive experiments on the open-loop nuScenes datasets and closed-loop Bench2Drive benchmark show that FocalAD outperforms state-of-the-art methods in planning accuracy and safety. Moreover, on the more challenging adversarial Adv-nuScenes dataset, FocalAD demonstrates strong robustness in high-risk interactive scenarios, highlighting the effectiveness of focusing on decision-critical local agents. Future work will explore generative planning frameworks and trajectory refinement strategies to further improve trajectory diversity and enhance planning safety.

\backmatter

\section*{Declarations}

\subsection*{Funding}

This work is supported by the National Key Research and Development Program of China (Grant No. 2022YFB2503000).

\subsection*{Conflict of interest}

On behalf of all the authors, the corresponding author states that there is no conflict of interest.





\bibliography{sn-bibliography}

\end{document}